\documentclass[runningheads]{llncs}
\usepackage[T1]{fontenc}
\usepackage{graphicx}
\usepackage{caption}
\usepackage{amsmath}
\usepackage{cleveref}
\usepackage{algorithm}
\usepackage{algpseudocode}
\usepackage{float}
\usepackage{array}
\usepackage{diagbox}

\usepackage{etoolbox} 
\makeatletter
\patchcmd{\ps@headings}{\rlap{\thepage}}{}{}{}
\patchcmd{\ps@headings}{\llap{\thepage}}{}{}{}
\makeatother
\usepackage[misc]{ifsym}

\begin{document}

\title{Regularized Multi-LLMs Collaboration for Enhanced Score-based Causal Discovery}
\titlerunning{}
%
\author{Xiaoxuan Li\inst{1}\textsuperscript{\Letter} \and
Yao Liu\inst{2,3} \and
Ruoyu Wang\inst{1}\and 
Lina Yao\inst{1,4}
}
\authorrunning{ }

\institute{
University of New South Wales, Sydney, Australia\\
\email{x.l.li@student.unsw.edu.au  ruoyu.wang5@unsw.edu.au}
\and
School of Computer Science and Engineering, Northeastern University, Shenyang, China, 110169\\
\email{liuyao@cse.neu.edu.cn}\\
\and
School of Computing, Macquarie University, Sydney, Australia\\
\and
CSIRO's Data61\\
\email{lina.yao@data61.csiro.au}
}

\maketitle              
%

\begin{abstract}

As the significance of understanding the cause-and-effect relationships among variables increases in the development of modern systems and algorithms, learning causality from observational data has become a preferred and efficient approach over conducting randomized control trials. However, purely observational data could be insufficient to reconstruct the true causal graph. Consequently, many researchers tried to utilise some form of prior knowledge to improve causal discovery process. In this context, the impressive capabilities of large language models (LLMs) have emerged as a promising alternative to the costly acquisition of prior expert knowledge. In this work, we further explore the potential of using LLMs to enhance causal discovery approaches, particularly focusing on score-based methods, and we propose a general framework to utilise the capacity of not only one but multiple LLMs to augment the discovery process.

\end{abstract}

\section{Introduction}

\noindent Causal discovery endeavours to uncover the cause-and-effect relationships among variables, revealing how changes in one can influence others\cite{JUDEA09}. 
Understanding causality becomes increasingly important in many fields, such as economics\cite{JUDEA09} and biology\cite{KODD05}. While conducting randomized control trials (RCTs) is the golden rule for testing causality, the execution of RCTs can be prohibitively costly, time-consuming, and ethically problematic in various domains. To avoid these inevitable problems, researchers seek to reconstruct causal relationships only relying on observational data without using RCTs. However, learning causality from observational data makes it difficult to distinguish the correct causal graph from a set of similar distributions\cite{ILJU08}. Consequently, one popular research direction is to use some other supplementary information to augment or guide the causal discovery process, thereby enhancing the reliability of outcomes\cite{CHRT23}. 
Supplementary information such as experts' domain knowledge and findings from RCTs could not only improve the accuracy of existing causal discovery methods but also reduce the search space hence accelerating the discovery process.

\begin{figure}[t]
    \centering
    \includegraphics[width = 11cm]{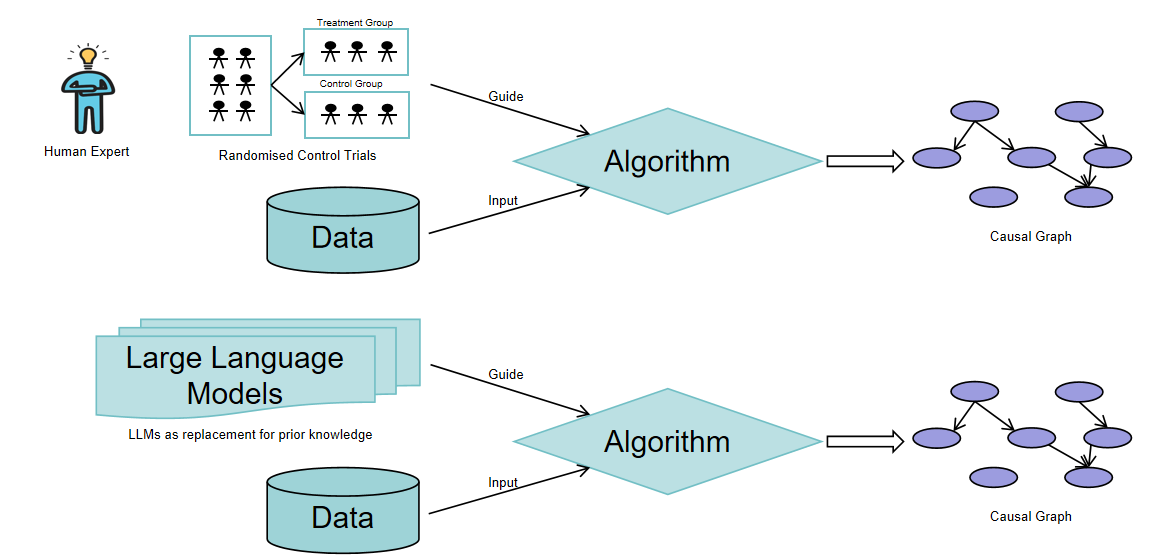}
    \caption{LLMs serve as potential substitutes for prior knowledge.}
    \label{fig:1}
\end{figure}

An inherent challenge in using prior knowledge is its unavailability in many cases, obtaining it could be time-consuming, expensive, or even impossible. The recent emergence of Large Language Models (LLMs) provides a potential solution to this challenge\cite{LOSP23,WZDK22}. LLMs have demonstrated their incredible capabilities in NLP tasks like text generation\cite{LTZW21} and sentiment analysis\cite{ZYZB23}. In causal discovery, LLMs offer the possibility of recovering cause-and-effect relationships using the description or only the name of variables, which could be treated as an alternative to costly expert knowledge and RCTs, Figure \ref{fig:1}. Numerous studies\cite{LOSP23,WZDK22} have investigated the efficacy of LLMs in common-sense causality discovery, drawing upon the vast knowledge upon which they are trained.

In this work, we delve into the capacity of LLMs to infer causal relationships. To our best knowledge, all the existing works harness the power of a single LLM to improve the causal discovery approach. In contrast, we proposed a framework to integrate multiple LLM agent results within score-based methodologies. The information provided by a solitary LLM is often limited and sparse, while our approach, combining multiple LLMs, yields more reliable and accurate results. To a certain extent, our framework improved the ability of LLM to infer causality, and hence further improved the accuracy and efficiency of the existing algorithms. The main contributions of this paper:
\begin{itemize}
    \item We introduce a novel framework that incorporates information from LLMs as an alternative to traditional prior knowledge within score-based methods for causal discovery tasks;
    \item We improve the accuracy of LLM results by combining information from multiple LLMs using a weighted sum, demonstrating this approach on existing score-based algorithms GES and NOTEARS;
    \item We validate the effectiveness of our framework through experiments on two score-based methods (GES, NOTEARS) and a reinforcement learning based causal discovery algorithm (KCRL), showing enhanced performance and superior results.
\end{itemize}

\section{Related Works}

\noindent\textbf{Causal Discovery} Methods for causal structure learning typically fall into two categories: constraint-based methods and score-based methods. The constraint-based methods reconstruct the causal graph by examining the properties of conditional independencies, such as PC\cite{JUPE22} and FCI\cite{SPGS00} algorithm; the score-based methods evaluate various estimated causal graphs by a predefined score function such as Bayesian Information Criterion(BIC)\cite{MCHE97} and BDe(u)\cite{HGMC95}, seeking to find the graph that minimises the objective score and satisfies the acyclicity constraint\cite{BDGP13,CHIC02,HAGA22,ZARX18}. A prominent score-based algorithm is Greedy Equivalence Search(GES)\cite{CHIC02} which utilises a greedy method to guide the search process. However, score-based methods encounter scalability problems due to the super-exponential growth of the search space with respect to the number of nodes. NOTEARS\cite{ZARX18} firstly introduced a smooth characterization for the acyclicity hence converting the combinatorial optimization problem to a continuous optimization problem, enabling the utilization of existing numerical and gradient-based methods for solving the problem.


\noindent\textbf{Causal Discovery with Prior Knowledge} Numerous researchers have tried to leverage the benefit of prior knowledge.  Wang et al. \cite{WHYY20} studied a specific problem with Type II diabetes and proposed a PKCL algorithm. Hasan and Gani \cite{HAGA22} introduced a framework called KCRL to leverage prior knowledge into the reinforcement learning causal discovery method proposed by Zhu et al. \cite{ZHNC20}. Hasan and Gani \cite{HAGA23} leveraged prior knowledge within the context of GES and introduced a novel method termed KGS. Chowdhury et al. \cite{CHRT23} integrated experts' knowledge into NOTEARS\cite{ZARX18} as some additional constraints.

\noindent\textbf{Causal Discovery with LLMs} The emergence of LLMs also gives a potential direction for improving causal discovery methods. Ban et al. \cite{BCWC23} introduced an innovative framework to integrate knowledge-based LLM with data-driven causal structure learning. Long et al. \cite{LOSP23} studied the capacity of LLMs to build causal graphs and discussed their potential to complement causal graph development. Long et al. \cite{LPZS23} treat the LLMs as an imperfect expert and incorporate this imperfect knowledge into current causal discovery mechanisms.


\section{Methodology}

In this section, we first define the causal discovery task. Then, we elaborate on our framework, including the acquisition of LLMs, and the combination and imposition of LLM-derived information into the existing score function as an additional penalty term. Furthermore, we illustrate two examples demonstrating the integration of the penalty term into specific algorithms, GES and NOTEARS.

\subsection{Problem Definition}

\noindent The causal structure learning problem is defined as the following: Given the observational dataset $X \in R^{n\times d}$, consisting of $n$ data and each data contains $d$ features. The task of causal discovery is to learn a Directed Acyclic Graph (DAG) $G(V, E)$ where each node $v\in V$ corresponding to a feature and the presence of an edge $e = (i, j)\in E$ indicates that feature $x_i$ is a direct cause of feature $x_j$. To retrieve the true causal relation, the distribution of the dataset $P(X)$ needs to be \textit{Markov} to the graph $G$ we learnt, which means $p(x) = \prod^{d}_{i=1}p_{i}(x_i~|~pa(x_i))$ where $pa(x_i)$ represents the parent set of $x_i$ in the graph $G$.

The \textbf{Score-based} methods formulate the problem as a combinatorial optimization problem: Given a DAG $G$ and a score function $S$, the score of the graph $S(G, X)$ indicates how good the graph is corresponding to the data $X$ (normally lower the better). Hence, the problem becomes an optimization problem:
\begin{align*}
    \min&_{G}\; S(G, X)\\
    \text{subject}&\text{ to }\; G\in \text{DAGs} 
\end{align*}

\begin{figure}[t]
    \centering
    \includegraphics[width = 9cm]{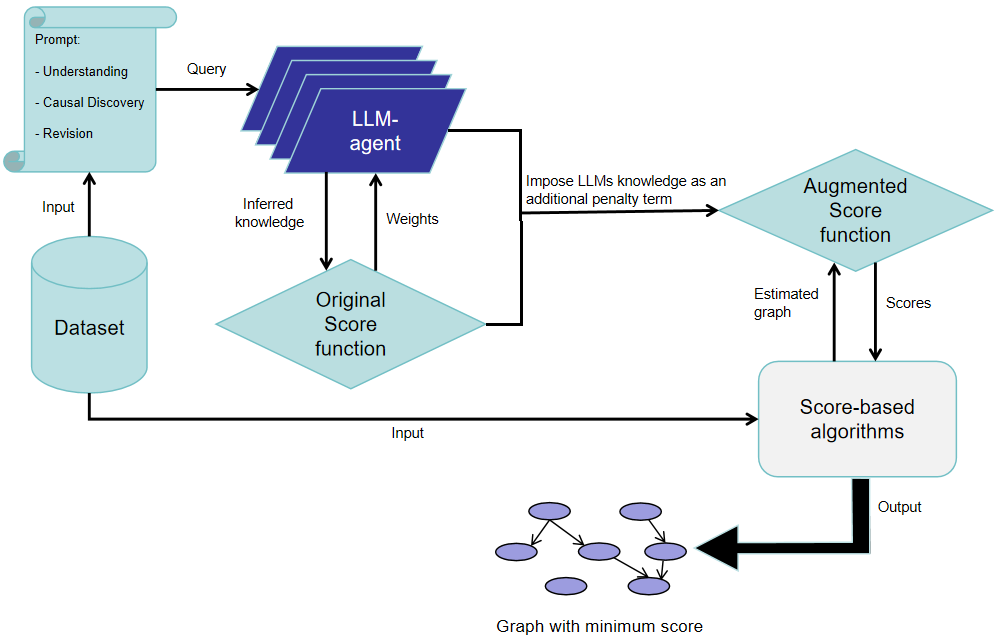}
    \caption{Overview of our framework.}
    \label{fig:2}
\end{figure}

Our framework seeks to integrate the knowledge derived from LLMs' knowledge into the learning process of any score-based causal discovery methods, as shown in Figure \ref{fig:2}. This process involves two primary stages: Initially, we interact with multiple LLMs to inquire about the dataset and establish causal relationships by combining the information from different LLMs; after getting the result, we incorporate the findings from LLMs into our causal discovery scoring function by introducing an additional penalty term. 

\subsection{Querying the LLMs}

Initially, information retrieval is conducted through querying the LLMs. To make the information useful and more reliable, we need to carefully design our prompt to fully harness the capabilities of LLMs. Ban et al. \cite{BCWC23} presented a prompt technique that consists of three stages: \textbf{U}nderstanding, \textbf{C}ausal Discovery and \textbf{R}evision. In the initial phase, the dataset is introduced to the LLM to facilitate comprehension of each variable by giving the LLM each variable's name and possible values; then, the LLM is asked to give the cause-and-effect relationships between variables based on this comprehension; and finally, a self-checking mechanism is employed to identify potential inaccuracies in the generated statements. 
Hence, the final result $K$ is a sequential outcome obtained by performing the three stages in sequence and can be expressed as $K = \textbf{U}\circ\textbf{C}\circ\textbf{R}$ where $\circ$ represents the sequential operation.



\subsection{Imposing Multi-LLMs' Collective Knowledge}

After the initial stage, the outcomes yielded by LLMs are obtained. However, note that these results are not infallible, and different LLMs would produce results of various quality. We proposed a framework designed to integrate diverse outcomes from multiple LLMs and impose any score-based method.

For any score-based approach, the fundamental objective is to find a DAG $G$ that optimises the score across all feasible graphs. Based on any existing score function $S_{\text{score}}(G)$, we introduce an additional penalty term to encapsulate the disparity between the graph and the outcomes produced by LLMs with a hyper-parameter $\lambda$ to reflect our degree of confidence in the LLM outcomes. The augmented score function is:
\[
    S(G) = S_{\text{score}}(G) + \lambda\times P(G)
\]

To compute the penalty term, we primarily explore two methods: $l_1$-penalty and $l_2$-penalty to suit different score-based methodologies and scenarios. Let $\hat{M}$ be the adjacency matrix corresponding to the estimated graph $\hat{G}$ outputted by the original score-based method, and $\hat{M}_{LLM}$ be the adjacency matrix corresponds to the predicted graph $\hat{G}_{LLM}$ generated by our LLMs from the first stage. The $l_1$-penalty quantifies the absolute difference between the estimated graph and the LLM results, i.e. $||\hat{M} - \hat{M}_{LLM}||$, this is equivalent to assessing the discrepancy between the estimated graph and the LLM outcomes. The $l_2$-penalty computes the squared difference between the estimated graph and the LLM results, represented as $||\hat{M} - \hat{M}_{LLM}||^2$.

As we employ multiple LLMs, we initially assess the quality of each LLM result using the identical score function and dataset to get a score denoted as $S_{score}(G_{LLM})$ for each LLM. Subsequently, we normalize them to ensure their summation equals one, obtaining a weight $\mu$ for each LLM model. Therefore, the penalty term is: 
\[
P_{l_1}(G) = \sum\limits_{\text{for each model}} \mu_{model} \times||\hat{M} - \hat{M}_{model}||
\]
\[
P_{l_2}(G) = \sum\limits_{\text{for each model}} \mu_{model} \times||\hat{M} - \hat{M}_{model}||^2
\]

Our framework is designed to suit any score-based methods. For demonstration purposes, we selected GES \cite{CHIC02} and NOTEARS \cite{ZARX18}. Many works are built upon these two works, hence, we chose them to illustrate the effectiveness of our framework. In the following, we have created two frameworks that integrate multiple LLMs into these methods, while also establishing a robust theoretical foundation. Both the GES and NOTEARS commonly use BIC\cite{MCHE97} as the score function:
\[
    \text{BIC} = -2 * loglikelihood + d * log(n)
\] where $d$ is the number of variables and $n$ is the size of training dataset.

\noindent\textbf{Muti-LLM Enhanced GES} The GES is a well-known score-based method that employs a greedy approach to guide the search process. GES requires that the score function is decomposable which could be written as the sum score among each node\cite{CHIC02}:
\[
    S(\hat{M}) = \sum_{i=1}^d s(x_i)
\]

We want to prove that our $l_1$-penalty and $l_2$-penalty are both decomposable and, hence could be added to the score function used in GES. 

The proof for $l_1$-penalty: 
\[
    P_{l_1}(\hat{M}) = ||\hat{M} - \hat{M}_{LLM}|| =\sum_{i=1}^d\sum_{j=1}^d |~\hat{M}[i][j] - \hat{M}_{LLM}[i][j]~| = \sum_{i=1}^d p(x_i)
\] where $p(x_i) = \sum_{j=1}^d |~\hat{M}[i][j] - \hat{M}_{LLM}[i][j]~|$. \\

The proof for $l_2$-penalty:
\[
    P_{l_2}(\hat{M}) = ||\hat{M} - \hat{M}_{LLM}||^2 =\sum_{i=1}^d\sum_{j=1}^d |~\hat{M}[i][j] - \hat{M}_{LLM}[i][j]~|^2 = \sum_{i=1}^d p(x_i)
\] where $p(x_i) = \sum_{j=1}^d |~\hat{M}[i][j] - \hat{M}_{LLM}[i][j]~|^2$.

We prove that both the penalties are decomposable, therefore, we can impose our LLM result penalty into GES without modifying the overall greedy mechanism.

\noindent\textbf{Multi-LLM Enhanced NOTEARS} NOTEARS\cite{ZARX18} proposed a novel equivalent acyclicity constraint to make the original combinatorial optimization problem to a continuous optimization problem by introducing a smooth characterisation of acyclicity constraint, consequently, this enables the application of gradient-based methods for solving the continuous problem:
\begin{align*}
    \min&_{M}\; F(M, X)\\
    \text{subject}&\text{ to }\; h(M) = 0
\end{align*} where $h(M) = 0$ indicates the graph $G$ induced by $M$ is acyclic and $h$ is differentiable, and $F$ is a continuous version of the score function.

To integrate our LLM-enhanced framework, we add the $l_2$-penalty term to the objective function $F$ in the NOTEARS, therefore, we need to prove that the $l_2$-penalty is differentiable and calculate the derivative. Given the $l_2$-penalty $P_{l_2}(\hat{M}) = \sum\limits_{\text{model}} \mu_{model}||\hat{M} - \hat{M}_{model}||^2$ and we could have the derivative of the $l_2$-penalty is:
\[
    \nabla P_{l_2}(\hat{M}) = \sum\limits_{\text{model}} - 2\times\mu_{model} (\hat{M} - \hat{M}_{model})
\]

Therefore, we add the $l_2$-penalty into the score function and the new objective function is:
\[
    F_{new}(\hat{M}) = F(\hat{M}) + \lambda\times P_{l_2}(\hat{M}) = F(\hat{M}) + \lambda\times\sum\limits_{\text{model}}\mu_{model} ||\hat{M} - \hat{M}_{model}||^2
\] with derivative:
\[
    \nabla F_{new}(\hat{M}) = \nabla F(\hat{M}) - 2\lambda\times\sum\limits_{\text{model}}\mu_{model} (\hat{M} - \hat{M}_{model}) 
\] hence, we could use the same Augmented Lagrangian method used in \cite{ZARX18} to solve the augmented continuous optimization problem.

\section{Experiments}
In this section, We evaluate the efficacy of our framework through experimentation across various datasets and compare the results generated by several score-based methods with those from the same methods enhanced by our LLM framework.

\subsection{Datasets and Evaluation Metrics}




To ensure the comprehensiveness of our experimental evaluation, we employed a diverse set of benchmark datasets, such as \textbf{LUCAS}, \textbf{Asia}, \textbf{Earthquake}, \textbf{Child} and \textbf{SACHS}, including both synthetic and real-world data. 


We mainly assess our framework on three evaluation metrics: False Discovery Rate (\textbf{FDR}) which calculates the proportion of estimated edges that are false, with lower values indicating better performance; True Positive Rate (\textbf{TPR}) which measures the likelihood of correctly identifying true edges within the estimated graph, with higher values indicating better performance;
\[
    fdr = \text{number of wrong edges discovered } / \text{ number of total edges discovered}
\]
\[
    tpr = \text{number of correct edges discovered } / \text{ number of total correct edges}
\]  
And Structural Hamming Distance (\textbf{SHD}) which quantifies the number of edge insertions, deletions, or flips necessary to transform the estimated graph into the true causal graph, with lower values indicating better performance.

\subsection{Results}


\noindent We delve into the experiments conducted and the resultant findings. For the LLMs, we utilized GPT-3.5, GPT-4, and Gemini to gather information about each dataset. 

\begin{figure}[h]
    \centering
    \includegraphics[width = 10.5cm]{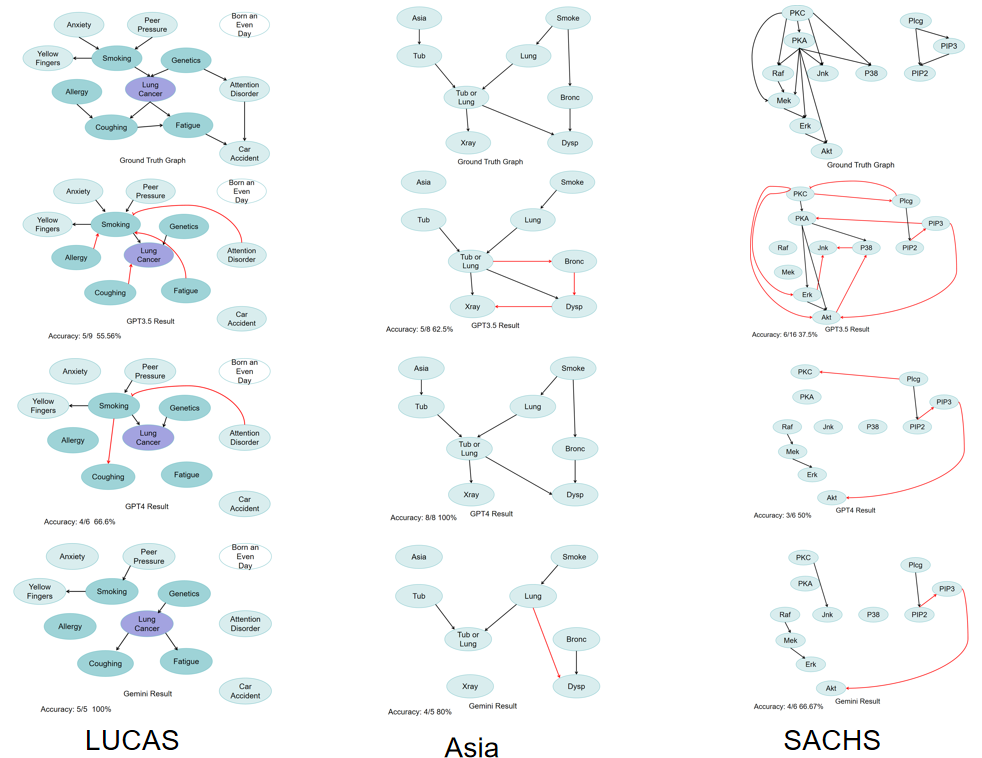}
    \caption{Information retrieved from GPT3.5, GPT4 and Gemini.}
    \label{fig:3}
\end{figure}

\noindent\textbf{Case Study} We analyze the efficacy of the information provided by each LLM on three datasets-LUCAS, Asia and SACHS. We queried GPT-3.5, GPT-4, and Gemini regarding the three datasets, and obtained the results, depicted in Figure \ref{fig:3} where black edges indicate correct predictions and red edges indicate incorrect predictions.

For the LUCAS dataset, Gemini yielded the most favourable outcome, correctly predicting all 5 edges. GPT-4 predicted 4 edges accurately but also generated 2 incorrect edges. Conversely, GPT-3.5 exhibited the poorest performance, predicting 5 correct edges but with 4 incorrect ones. Additionally, it is evident from the figure that while Gemini achieved the highest accuracy, both GPT-4 and GPT-3.5 still managed to predict some correct edges not identified by Gemini.

In the case of the Asia dataset, the results reveal that GPT-4 accurately predicted all 8 edges without generating any incorrect edges. This emphasises the remarkable capability of LLMs in causality discovery tasks. Contrastingly, GPT-3.5 predicted 5 correct edges but with 3 incorrect edges, while Gemini predicted 4 correct edges but with 1 incorrect edge. Despite these variations in performance, the fact that GPT-4 achieved perfect accuracy further highlights the potential of LLMs in uncovering causal relationships within datasets.

For the SACHS dataset, GPT-3.5 provided numerous causal relationships, yet a considerable portion were incorrect. On the other hand, both GPT-4 and Gemini tended to offer fewer statements, resulting in higher accuracy. 

The results underscore a notable observation: employing the same prompt technique yields varying information quality from different LLMs. Even LLMs with lower accuracy may predict certain correct edges not identified by LLMs with higher accuracy. This reaffirms our intuition regarding the integration of multiple LLMs. By doing so, we can access additional information that potentially exhibits higher quality, enriching our understanding and insights from the data.

\begin{table}[h!]

\renewcommand{\arraystretch}{0.9}
\newcolumntype{C}[1]{>{\centering\arraybackslash}m{#1}}

\caption{Experiment Results.}\label{tab1}

\begin{center}

\centerline{
\begin{tabular}{| c c | C{1.6cm} | C{1.6cm} !{\vrule width 2pt} C{1.6cm} | C{1.6cm} !{\vrule width 2pt} C{1.6cm} | C{1.6cm} |}
\hline
\multicolumn{2}{|c|}{\diagbox{Datasets}{Methods}} & GES & Enhanced GES & NOTEARS & Enhanced NOTEARS & KCRL & Enhanced KCRL\\
\hline
               & SHD & 1 & 1 & 7 & 7 & 9 & \textbf{6} \\
\textbf{LUCAS} & TPR & 1.0 & 1.0 & 0.42 & 0.42 & 0.41 & \textbf{0.5}  \\
               & FDR & 0.077 & 0.077 & 0.44 & 0.44 & 0.5 & \textbf{0.0} \\
\hline
               & SHD & 3 & 4 & 4 & 5 & 4 & \textbf{2} \\
\textbf{Asia}  & TPR & 0.875 & \textbf{1.0} & 0.5 & 0.375 & 0.5 & \textbf{0.75}  \\
               & FDR & 0.22 & 0.33 & 0.33 & \textbf{0.25} & 0.2 & \textbf{0.0}  \\
\hline
                    & SHD & 5 & 5 & 4 & \textbf{2} & 3 & \textbf{1} \\
\textbf{Earthquake} & TPR & 0.5 & 0.5 & 0.25 & \textbf{0.75} & 0.5 & \textbf{0.75}  \\
                    & FDR & 0.66 & 0.66 & 0.66 & \textbf{0.5} & 0.33 & \textbf{0.0}\\
\hline
               & SHD & 17 & \textbf{16} & 12 & \textbf{11} & 13 & 14  \\
\textbf{SACHS} & TPR & 0.35 & 0.35 & 0.35 & \textbf{0.41} & 0.23 & \textbf{0.29}  \\
               & FDR & 0.57 & \textbf{0.54} & 0.57 & \textbf{0.53} & 0.63 & \textbf{0.61}  \\
\hline
               & SHD & 23 & 26 & 18 & \textbf{16} & 20 & 21  \\
\textbf{Child} & TPR & 0.68 & 0.68 & 0.32 & \textbf{0.36} & 0.44 & 0.44  \\
               & FDR & 0.60 & 0.62 & 0.6 & \textbf{0.52} & 0.57 & \textbf{0.56}  \\
\hline
\end{tabular}
}
\end{center}
\end{table}

\noindent\textbf{Experiment Results Analysis} In addition to GES and NOTEARS, we also conducted experiments on KCRL to evaluate our framework. Rather than utilizing precise prior knowledge as KCRL did, we used our LLM prior knowledge instead to demonstrate the effectiveness of LLMs on causal discovery tasks. The experimental results for the GES, NOTEARS and KCRL algorithms, both with and without our framework, are summarized in Table \ref{tab1}.

From the table, we observe that the LLM enhancement leads to improvements across all datasets. For GES, in the Asia dataset, the TPR increases to 1, indicating that the algorithm can predict all the correct edges in the ground truth graph with the help of LLM. Despite this, SHD and FDR also increase, suggesting the algorithm predicts more spurious edges. In the SACHS dataset, we see pure improvements in SHD and FDR, while TPR remains the same as the original algorithm.

For NOTEARS, there is no performance change on the LUCAS dataset. On the Asia dataset, FDR is reduced, but SHD is higher, and TPR is lower in the LLM-enhanced version. Significant improvements are noted in the Earthquake, SACHS, and Child datasets, where the enhanced version outperforms the original in all three metrics.

For KCRL, we observe substantial improvements. The LLM-enhanced version outperforms the original algorithm in all three metrics on the LUCAS, Asia, and Earthquake datasets, even reducing FDR to 0 in these cases. On the SACHS and Child datasets, improvements are noted across several metrics.

\subsection{Discussion}
Throughout the experiments, we encountered challenges due to the quality of LLM results, which significantly limited the effectiveness of our approach. Despite incorporating a weight parameter to signify our confidence in the LLM results, tuning this parameter remains a complex task. Additionally, our current approach employs a weighted sum to combine information of varying quality, yet exploring more sophisticated methods for integrating this information could yield superior results, thereby enhancing the performance of original score-based methods further. Moreover, the incorporation of additional information into existing score functions warrants further investigation. Given the non-convex nature of the problem, conventional score-based methods often encounter challenges in escaping local minima. Designing improved score functions holds the potential to guide algorithms away from local minima towards global optimal solutions.

\section{Conclusion}
In this study, we introduce a novel framework aimed at integrating the capabilities of multiple LLMs into the score-based causal discovery methodology. In contrast to conventional methods guided by prior knowledge, our approach circumvents the need for potentially costly acquisition of expert knowledge or randomized control trials by leveraging the capacity of LLMs. We incorporate LLM information as an additional penalty term into existing score functions, thereby prompting the original methodology to account for this supplementary information. Through a series of diverse experiments, we demonstrate the efficacy of our framework in enhancing existing score-based algorithms.

%
%
%

\bibliographystyle{splncs04}
\bibliography{a-paper/refs}

\begin{thebibliography}{10}
\providecommand{\url}[1]{\texttt{#1}}
\providecommand{\urlprefix}{URL }
\providecommand{\doi}[1]{https://doi.org/#1}

\bibitem{BDGP13}
Alonso{-}Barba, J.I., delaOssa, L., G{\'{a}}mez, J.A., Puerta, J.M.: Scaling up the greedy equivalence search algorithm by constraining the search space of equivalence classes. Int. J. Approx. Reason.  (2013)

\bibitem{BCWC23}
Ban, T., Chen, L., Wang, X., Chen, H.: From query tools to causal architects: Harnessing large language models for advanced causal discovery from data. CoRR  (2023)

\bibitem{CHIC02}
Chickering, D.M.: Optimal structure identification with greedy search. J. Mach. Learn. Res.  (2002)

\bibitem{MCHE97}
Chickering, D.M., Heckerman, D.: Efficient approximations for the marginal likelihood of bayesian networks with hidden variables. Mach. Learn.  (1997)

\bibitem{CHRT23}
Chowdhury, J., Rashid, R., Terejanu, G.: Evaluation of induced expert knowledge in causal structure learning by {NOTEARS}. In: Proceedings of the 12th International Conference on Pattern Recognition Applications and Methods, {ICPRAM} 2023. {SCITEPRESS} (2023)

\bibitem{HAGA22}
Hasan, U., Gani, M.O.: {KCRL:} {A} prior knowledge based causal discovery framework with reinforcement learning. In: Proceedings of the Machine Learning for Healthcare Conference, {MLHC} 2022. Proceedings of Machine Learning Research, {PMLR} (2022)

\bibitem{HAGA23}
Hasan, U., Gani, M.O.: {KGS:} causal discovery using knowledge-guided greedy equivalence search. CoRR  (2023)

\bibitem{HGMC95}
Heckerman, D., Geiger, D., Chickering, D.M.: Learning bayesian networks: The combination of knowledge and statistical data. Mach. Learn.  (1995)

\bibitem{LTZW21}
Li, J., Tang, T., Zhao, W.X., Wen, J.: Pretrained language models for text generation: {A} survey. CoRR  (2021)

\bibitem{LPZS23}
Long, S., Pich{\'{e}}, A., Zantedeschi, V., Schuster, T., Drouin, A.: Causal discovery with language models as imperfect experts. CoRR  (2023)

\bibitem{LOSP23}
Long, S., Schuster, T., Pich{\'{e}}, A.: Can large language models build causal graphs? CoRR  (2023)

\bibitem{JUDEA09}
Pearl, J.: Causality: Models, Reasoning and Inference. Cambridge University Press (2009)

\bibitem{JUPE22}
Pearl, J.: Causality 2002-2020 - introduction. In: Probabilistic and Causal Inference: The Works of Judea Pearl. {ACM} Books, {ACM} (2022)

\bibitem{KODD05}
Sachs, K., Perez, O., Pe'er, D., Lauffenburger, D.A., Nolan, G.P.: Causal protein-signaling networks derived from multiparameter single-cell data. Science  (2005)

\bibitem{ILJU08}
Shpitser, I., Pearl, J.: Complete identification methods for the causal hierarchy  (2008)

\bibitem{SPGS00}
Spirtes, P., Glymour, C., Scheines, R.: Causation, Prediction, and Search, Second Edition. Adaptive computation and machine learning, {MIT} Press (2000)

\bibitem{WHYY20}
Wang, W., Hu, G., Yuan, B., Ye, S., Chen, C., Cui, Y., Zhang, X., Qian, L.: Prior-knowledge-driven local causal structure learning and its application on causal discovery between type 2 diabetes and bone mineral density. {IEEE} Access  (2020)

\bibitem{WZDK22}
Willig, M., Zecevic, M., Dhami, D.S., Kersting, K.: Can foundation models talk causality? CoRR  (2022)

\bibitem{ZYZB23}
Zhang, B., Yang, H., Zhou, T., Babar, A., Liu, X.: Enhancing financial sentiment analysis via retrieval augmented large language models. In: 4th {ACM} International Conference on {AI} in Finance, {ICAIF} 2023. {ACM} (2023)

\bibitem{ZARX18}
Zheng, X., Aragam, B., Ravikumar, P., Xing, E.P.: Dags with {NO} {TEARS:} continuous optimization for structure learning. In: Advances in Neural Information Processing Systems 31: Annual Conference on Neural Information Processing Systems 2018 (2018)

\bibitem{ZHNC20}
Zhu, S., Ng, I., Chen, Z.: Causal discovery with reinforcement learning. In: 8th International Conference on Learning Representations, {ICLR} 2020. OpenReview.net (2020)

\end{thebibliography}

\end{document}